\documentclass[11pt]{article}
\usepackage{authblk}

\usepackage{coling2020}
\usepackage[utf8]{inputenc}
\usepackage[russian,english]{babel}
\usepackage{times}
\usepackage{url}
\usepackage{latexsym}
\usepackage{graphicx}
\usepackage[table,xcdraw]{xcolor}
\usepackage{multirow}
\usepackage{booktabs}
\usepackage{enumitem}

\usepackage{subfig}
\DeclareRobustCommand{\cyrins}[1]{%
  \begingroup\fontfamily{erewhon-TLF}%
  \foreignlanguage{russian}{#1}%
  \endgroup
}
\colingfinalcopy 

\title{Studying Taxonomy Enrichment on Diachronic WordNet Versions} 

\author[$\ddag$]{\textbf{Irina Nikishina}}
\author[$\ddag$]{\textbf{Alexander Panchenko}}
\author[$\ddag$]{\textbf{Varvara Logacheva}}
\author[$\dag$]{\textbf{Natalia Loukachevitch}}

\affil[$\ddag$]{Skolkovo Institute of Science and Technology, Moscow, Russia}
\affil[$\dag$]{Research Computing Center, Lomonosov Moscow State University, Moscow, Russia}
\affil[ ]{\tt \{Irina.Nikishina,A.Panchenko,V.Logacheva\}@skoltech.ru}
\affil[ ]{\tt louk\_nat@mail.ru}

\date{}

\begin{document}
\maketitle
\begin{abstract}
  Ontologies, taxonomies, and thesauri are used in many NLP tasks. 
  However, most studies are focused on the creation of these lexical resources rather than the maintenance of the existing ones. 
  Thus, we address the problem of \textit{taxonomy enrichment}.
  We explore the possibilities of taxonomy extension in a resource-poor setting and present methods which are applicable to a large number of languages. We create novel English and Russian datasets for training and evaluating taxonomy enrichment models and describe a technique of creating such datasets for other languages. 
\end{abstract}

\section{Introduction}
\blfootnote{
    %
    %
    %
    %
    \hspace{-0.65cm}  
    This work is licensed under a Creative Commons 
    Attribution 4.0 International Licence.
    Licence details:
    \url{http://creativecommons.org/licenses/by/4.0/}.
    
    %
}
Nowadays, construction and maintenance of lexical resources (ontologies, knowledge bases, thesauri) have become essential for the NLP community.
In particular, enriching the most acknowledged lexical databases like WordNet  \cite{miller1998wordnet} and its variants for almost 50 languages\footnote{According to \url{http://globalwordnet.org/resources/wordnets-in-the-world}} 
or collaboratively created lexical resources such as Wiktionary
is crucial. 
Resources of this kind are widely used in multiple NLP tasks: Word Sense Disambiguation, Entity Linking \cite{moro-navigli-2015-semeval}, Named Entity Recognition, Coreference Resolution \cite{ponzetto-strube-2006-exploiting}.

There already exist several initiatives on WordNet extension, for example, the Open English WordNet 
with thousands of new manually added entries or plWordNet \cite{maziarz-etal-2014-plwordnet} which includes a mapping to an enlarged Princeton WordNet. However, the manual annotation process is too costly: it is time-consuming and requires language or domain experts. 
On the other hand, automatically created datasets and resources usually lag in quality compared to manually labelled ones. Therefore, it would be beneficial to assist manual work by introducing automatic annotation systems to keep valuable lexical resources up-to-date. In this paper, we analyse the approaches to automatic enrichment of wordnets.

Formally, the goal of the Taxonomy Enrichment task is as follows: given words that are not included in a taxonomy (hereinafter \textit{orphan words}), we need to associate each word with the appropriate hypernyms from it. For example, given an input word ``duck'' we need to provide a list of the most probable hypernyms the word could be attached to, e.g. ``waterfowl'', ``bird''. A word may have multiple hypernyms. 


SemEval-2016 task 14 \cite{jurgens-pilehvar-2016-semeval} was the first effort to evaluate this task in a controlled environment, but it provided an unrealistic task scenario. Namely, the participants were given definitions of words to be added to taxonomy, which are often unavailable in real world. The majority of presented methods heavily depended on these definition. In contrast to that, we present a resource-poor scenario and solutions which conform to it. Our contributions are as follows:

\begin{itemize}[noitemsep]

\item We create datasets for studying diachronic evolution of wordnets for English and Russian, extending the monolingual setup of the RUSSE'2020 shared task \cite{nikishina2020taxonomy} with a larger Russian dataset and similar English versions.\footnote{Code \& datasets available at \url{https://github.com/skoltech-nlp/diachronic-wordnets}}

\item We present several new simple yet efficient methods for taxonomy enrichment. They are applicable for the majority of languages and are competitive compared to the state-of-the-art results for nouns.

\end{itemize}

\section{Related Work}
\label{sec:related}

The existing studies on the taxonomies can be divided into three groups. The first one addresses the Hypernym Discovery problem \cite{camacho-collados-etal-2018-semeval}: given a word and a text corpus, the task is to identify hypernyms in the text. However, in this task, the participants are not given any predefined taxonomy to rely on. The second group of works deals with the Taxonomy Induction problem \cite{bordea-etal-2015-semeval,bordea-etal-2016-semeval,velardi-etal-2013-ontolearn}, 
in other words, creation of a taxonomy from scratch. Finally, the third direction of research is the Taxonomy Enrichment task: the participants extend a given taxonomy with new words. Our methods tackle this task.

Unlike the former two groups, the latter garners less attention. Until recently, the only dataset for this task was created under the scope of SemEval-2016. It contained definitions for new words, so the majority of models solving this task used the definitions. For instance, \newcite{tanev-rotondi-2016-deftor} computed definition vector for the input word, comparing it with the vector of the candidate definitions from WordNet using cosine similarity. Another example is \textit{TALN} team \cite{espinosa-anke-etal-2016-taln} which also makes use of the definition by extracting noun and verb phrases for candidates generation.

This scenario may be unrealistic for manual annotation because annotators are writing a definition for a new word and adding new words to the taxonomy simultaneously. 
Having a list of candidates would not only speed up the annotation process but also identify the range of possible senses. Moreover, it is possible that not yet included words may have no definition in any other sources: they could be very rare (``apparatchik'', ``falanga''), relatively new (``selfie'', ``hashtag'') or come from a 
narrow domain (``vermiculite'').

Thus, following RUSSE-2020 shared task \cite{nikishina2020taxonomy} we stick to a more realistic scenario when we have no definitions of new words, but only examples of their usage. The organisers of the shared task provided a baseline and training and evaluation datasets based on RuWordNet~\cite{loukachevitch2016creating}. The task exploited words which were recently added to the latest release of RuWordNet and for which the hypernym synsets for the words were already identified by qualified annotators. The participants of the competition were asked to find synsets which could be used as hypernyms.  


The participants of this task mainly relied on vector representations of words and the intuition that words used in similar contexts have close meanings. They cast the task as a classification problem where words need to be assigned one or more hypernyms \cite{kunilovskaya2020russe} or ranked all hypernyms by suitability for a particular word \cite{dale2020russe}. They also used a range of additional resources, such as Wiktionary \cite{arefyev2020russe}, dictionaries, additional corpora.
Interestingly, only one of the well-performing models \cite{tikhomirov2020russe} used context-informed embeddings (BERT).


However, the best-performing model (denoted as \textit{Yuriy} in the workshop description paper) extensively used external tools such as online Machine Translation (MT) and search engines. This approach is difficult to replicate because their performance for different languages can vary significantly. Thus, we exclude features which relate to search engines and compute the model performance without them.

The same applies to pre-trained word embeddings to some extent, but in this case we know which data was used for their training and can make more informed assumptions about the downstream performance. Moreover, the embeddings are trained on unlabelled text corpora which exist in abundance for many languages, so training high-quality vector representations is an easier task than developing a well-performing MT or search engine.

Therefore, we would like to work out methods which do not depend on resources which exist only for a small number of well-resourced languages (e.g. semantic parsers or knowledge bases) or data which should be prepared specifically for the task (descriptions of new words). At the same time, we want our methods to benefit from the existing data (e.g. corpora, pre-trained embeddings, Wiktionary).



\section{Diachronic WordNet Datasets}
\label{sec:dataset}

We build two diachronic datasets: one for English, another one for Russian based respectively on Princeton WordNet \cite{miller1998wordnet} and RuWordNet taxonomies. Each dataset consists of a taxonomy and a set of novel words 
not included in the taxonomy.

\begin{table}[ht]
\centering
\begin{tabular}{lrrrrrr}
\toprule
\multicolumn{1}{c}{\multirow{2}{*}{\textbf{Taxonomy}}} & \multicolumn{2}{c}{\textbf{Synsets}}                  & \multicolumn{2}{c}{\textbf{Lemmas}}                   & \multicolumn{2}{c}{\textbf{New words}}                      \\ 
\multicolumn{1}{c}{}                                   & \multicolumn{1}{c}{Nouns} & \multicolumn{1}{c}{Verbs} & \multicolumn{1}{c}{Nouns} & \multicolumn{1}{c}{Verbs} &
\multicolumn{1}{c}{Nouns} & \multicolumn{1}{l}{Verbs} \\ \midrule
\textit{WordNet 1.6}                                   &             66 025              &              12 127             &             94 474              &       10 319                    &        -                   &     -    \\
\textit{WordNet 1.7}                                   &             75 804              &              13 214             &             109 195              &       11 088                    &        11 551  &     401        \\
\textit{WordNet 2.0}                                   &             79 689              &              13 508             &             114 648              &       11 306                    &        4 036 & 182 \\
\textit{WordNet 2.1}                                   &             81 426              &              13 650             &             117 097              &       11 488                    &       2 023 & 158  \\
\textit{WordNet 3.0}                                   &           82 115                &           13 767                &          117 798                 &        11 529                   &          678              &   33    \\ 
\bottomrule
\end{tabular}
\caption{Statistics of the English WordNet taxonomies used in this study.}
\label{tab:statistics}
\end{table}

\begin{table}[ht]
    \centering
\begin{tabular}{lrr}
\toprule
\textbf{Dataset} & \textbf{Nouns} &\textbf{Verbs} 
\\ \midrule
\textit{WordNet 1.6 - WordNet 3.0}     & 17 043  & 755 \\
\textit{WordNet 1.7 - WordNet 3.0}     & 6 161  & 362 \\
\textit{WordNet 2.0 - WordNet 3.0}     & 2 620  & 193 \\
\midrule
\textit{RuWordNet 1.0 - RuWordNet 2.0}     & 14 660 & 2 154  \\
\textit{RUSSE'2020}     & 2 288 & 525  \\
\bottomrule
\end{tabular}
    \caption{Statistics of the diachronic wordnet datasets used in this study.}
    \label{tab:dataset}
\end{table}

\subsection{English Dataset} We choose two versions of WordNet and then select words which appear only in a newer version. For each word, we get its hypernyms from the newer WordNet version and consider them as gold standard hypernyms. We add words to the dataset if only their hypernyms appear in both snippets. We do not consider adjectives and adverbs, because they often introduce abstract concepts and are difficult to interpret by context. Besides, the taxonomies for adjectives and adverbs are worse connected than those for nouns and verbs making the task more difficult.

In order to find the most suitable pairs of releases, we compute WordNet statistics (see Table \ref{tab:statistics}). New words demonstrate the difference between the current and the previous WordNet version. For example, it shows that the dataset generated by ``subtraction'' of WordNet 2.1 from WordNet 3.0 would be too small, they differ by $678$ nouns and $33$ verbs. Therefore, we create several datasets by skipping one or more WordNet versions.
The statistics for each dataset are provided in Table \ref{tab:dataset}.

As gold standard hypernyms, we use not only the immediate hypernyms of each lemma but also the second-order hypernyms: hypernyms of the hypernyms. We include them in order to make the evaluation less restricted. According to our empirical observations, the task of automatically identifying the exact hypernym might be too challenging, and finding the region where a word belongs (``parents'' and ``grandparents'') can already be considered a success.

\subsection{Russian Dataset}
Our method of dataset construction does not use any language-specific or database-specific features. We show how it can be transferred to other wordnets or taxonomies with timestamped releases for Russian. We create an analogous version to English  extending the dataset by \newcite{nikishina2020taxonomy} based on RuWordNet. 
%
The original dataset does not include short words ($<4$ symbols), diminutives, named entities and other constraints described in the shared task paper. We remove those constraints and present a non-restricted Russian dataset and a symmetrical English dataset from WordNet database (cf. Table~\ref{tab:dataset}).

\section{Taxonomy Enrichment Methods}
\label{sec:baselines}


Our method is based on the baseline distributional model from RUSSE-2020 shared task extended with ranking of synset candidates and use the information from Wiktionary and various types of embeddings. 

\subsection{Baseline}

According to \newcite{cai2018improving} and \newcite{aly-etal-2019-every}, co-hyponyms (words or phrases that share a hypernym) usually have similar contexts. On the other hand, the distributional hypothesis states that words that occur in similar context tend to have similar meanings \cite{harris1954distributional}. Our approach is built on the top of  the baseline method by~\newcite{nikishina2020taxonomy}. In this method, top $k=10$ nearest neighbours of the input word are taken from the pre-trained embedding model (according to the above considerations they should be co-hyponyms). Subsequently,  hypernyms  of those co-hyponyms are extracted from the taxonomy. These hypernyms can also be considered hypernyms of the input word. 

There is no one-to-one mapping between a word and a synset. On one hand, several hypernyms can belong to one synset, on the other hand, one word can occur in multiple synsets. Thus, all synsets associated with the list of extracted hypernyms are extracted. Then, vector representation of a synset is computed by averaging embeddings of all lemmas belonging to the synset. The model returns top $k$ closest synsets instead of top $k$ words. 
Despite its simplicity, this method turned out to be a strong baseline as it outperformed over half of the participating models.


\subsection{Ranking Extended Hypernyms List by Weighted Similarity}

This baseline has a shortcoming: it lacks sorting operation on the extracted candidates. The rank of synsets is defined only by the rank of a corresponding nearest neighbour.

We improve the described model by ranking the generated synset candidates. In addition to that, we extend the list of candidates with second-order hypernyms (hypernyms of each hypernym). The direct hypernyms of the word's nearest neighbours can be too specific, whereas second-order hypernyms are likely to be more abstract concepts, which the input word and its neighbours have in common. After forming a list of candidates, we score each of them using the following equation:
%
    $score_{h_{i}} = n \cdot sim(v_o, v_{h_{i}}),$
%
where $v_x$ is a vector representation of a word or a synset $x$, $h_i$ is a hypernym, $n$ is the number of occurrences of this hypernym in the merged list, $sim(v_o, v_{h_{i}})$ is the cosine similarity of the vector of the orphan word $o$ and hypernym vector $h_i$. By computing this score, we assume that the most frequent and the most similar candidates are the true hypernyms of the word. We sort the hypernyms by this score and return top $k$. 

\subsection{Features Extracted from Wiktionary}

One of the promising multilingual resources which the taxonomy enrichment models could benefit from 
is Wiktionary.
We choose it as Wiktionary is the only large web-based free content dictionary existing for $175$ languages, including English ($6,334,384$ entries) and Russian ($1,076,156$ entries). More importantly, each Wiktionary page usually comprises a definition and lists of hypernyms, hyponyms and synonyms, which could be useful for our task. 
We implement the following Wiktionary features:
\begin{itemize}[noitemsep]
    \item the candidate is present in the Wiktionary hypernyms list for the input word (binary feature),
    \item the candidate is present in the Wiktionary synonyms list (binary feature),
    \item the candidate is present in the Wiktionary definition (binary feature),
    \item average cosine similarity between the candidate and the Wiktionary hypernyms of the input word.
\end{itemize}

We do not use the definitions directly, as their texts are too noisy. They often include example usages of words which cannot be separated from the definitions and can distort their vector representations.


We extract lists of hypernym synset candidates using the baseline procedure and compute the 4 Wiktionary features for them. In addition to that, we use the score from the previous approach as a feature. To define the feature weights, we train a Logistic Regression model with L2 regularisation on a training dataset which we construct from the older (known) versions of WordNet. This dataset is constructed analogously to the datasets for evaluation which we described in Section \ref{sec:dataset} using all leaf synsets from the older WordNet. For each lemma of such synsets, we extract their gold standard hypernym synsets.  As a result, our dataset comprised $79,000$ positive and $79,000$ negative examples of \textit{word-candidate} pairs for both nouns and verbs for English dataset and $59,914$ and $59,914$ for Russian.

In order to understand the contribution of the Wiktionary features to the final score we compute the number of orphans encountered in Wiktionary (97\% to 100\%) and the number of orphans containing at least one hypernym in Wiktionary fields (2--18\% for ``hypernyms'' field, 1--2\% for ``synonyms'' field and 26--35\% in the definition). 



\subsection{Pre-trained Embeddings}

We test our methods on non-contextualised fastText \cite{bojanowski-etal-2017-enriching} and contextualised BERT \cite{devlin-etal-2019-bert} embeddings.
We choose fastText embeddings because pre-trained fastText models are easy to deploy, and do not require additional data or training for the out-of-vocabulary words. In this paper we use the fastText embeddings from the official website\footnote{\url{https://fasttext.cc/docs/en/crawl-vectors.html}} for both English and Russian, trained on Common Crawl from 2019 and Wikipedia CC including lexicon from the previous periods as well. 

While fastText embeddings can be generated for individual words, BERT requires a context for a word (i.e. a sentence containing it) to generate its embedding. For experiments with English datasets, we extract contexts from Wikipedia. For the experiments with Russian, we use a news corpus provided by the organisers of RUSSE'2020,\footnote{\url{https://github.com/dialogue-evaluation/taxonomy-enrichment}} which contains at least 50 occurrences for each word in the dataset. 


We use the pre-trained BERT-base model for English from \cite{devlin-etal-2019-bert}. For Russian, we utilize RuBERT model from \cite{kuratov2019adaptation}, which proved to outperform the Multilingual BERT from the original paper. To compute BERT embeddings for orphans and synsets, we extract sentences containing them from the corresponding corpora. If the words are absent in the corpora, we computed the average of lemmas without context for synsets and the embedding of the input word without context. We also averaged word-pieces for the OOV words. We lemmatise corpora with UDPipe \cite{udpipe:2017} to be able to find not only exact word matches but also their grammatical forms.
We rely on UDPipe as it supports many languages and shows reasonable performance on our data. In case of multiple occurrences of the same orphan, we average the retrieved contextualised embeddings. 

\section{Experiments}
\label{sec:evaluation}

\subsection{Evaluation Metric}

We consider the Taxonomy Enrichment task as a soft ranking problem and use Mean Average Precision (MAP) score for the quality measurement:

$$MAP = \frac{1}{N} \sum_{i=1}^{N} AP_{i}; AP_{i} = \frac{1}{M} \sum_{i}^{n} prec_{i} \times I[y_{i} = 1], $$

where $N$ and $M$ are the number of predicted and ground truth values, respectively, $prec_i$ is the fraction of ground truth values in the predictions from 1 to $i$, $y_i$ is the label of the $i$-th answer in the ranked list of predictions, and $I$ is the indicator function.

This metric is widely acknowledged in the Hypernym Discovery shared tasks, where systems are also evaluated over the top candidate hypernyms \cite{camacho-collados-etal-2018-semeval}. 
The MAP score takes into account the whole range of possible hypernyms and their rank in the candidate list.

However, 
the design of our dataset disagrees with MAP metric. As we described in Section \ref{sec:dataset}, the gold-standard hypernym list is extended with second-order hypernyms (parents of parents). This extension can distort MAP. If we consider all gold standard answers as compulsory for the maximum score, it means that we demand models to find both direct and second-order hypernyms. This disagrees with the original motivation of including second-order hypernyms to the gold standard --- it was intended to make the task easier by allowing a model to guess a direct \textit{or} a second-order hypernym.

On the other hand, if we decide that guessing \textit{any} synset from the gold standard yields the maximum MAP score, we will not be able to provide an adequate evaluation for words with multiple direct hypernyms. There exist two cases thereof:

\begin{enumerate}[noitemsep]
    \item the target word has two or more hypernyms which are co-hyponyms or one is a hypernym of the other --- this word has a single sense, but the annotator decided that multiple related hypernyms are needed to reflect all shades of the meaning,
    \item the target word has two or more hypernyms which are not directly connected in the taxonomy and neither are their hypernyms. This happens if:
    \begin{enumerate}[noitemsep]
        \item the word's sense is a composition of senses of its hypernyms, e.g. ``impeccability'' possesses two components of meaning: (``correctness'', ``propriety'') and  (``morality'', ``righteousness'');
        
        \item the word is polysemous and different hypernyms reflect different senses, e.g. ``pop-up'' is a book with three-dimensional pages (``book, publication'') and a baseball term (``fly, hit'').
        
    \end{enumerate}
\end{enumerate}

While the case 2a corresponds to a monosemous word and the case 2b indicates polysemy, this difference does not affect the evaluation process. We suggest that in both these cases in order to get the maximum MAP score a model should capture all the unrelated hypernyms which correspond to different components of sense. At the same time, we should bear in mind that guessing a direct hypernym or a second-order hypernym are equally good options. Therefore, following \newcite{nikishina2020taxonomy}, we evaluate our models with modified MAP. It transforms a list of gold standard hypernyms into a list of connected components. Each of these components includes hypernyms (both direct and second-order) which form a connected component in a taxonomy graph. (According to graph theory, connected component is a subgraph, in which there is a path between any two nodes.) Thus, in the case 1 we will have a single connected component, and a model should guess \textit{any} hypernym from it to get the maximum MAP score. In the cases 2a and 2b we will have multiple components, and a model should guess \textit{any} hypernym from \textit{each} of the components.



\begin{table}[]
\centering
\begin{tabular}{l|ll|ll}
\toprule
\multicolumn{1}{c|}{\multirow{2}{*}{\bf Method}} & \multicolumn{2}{c|}{\bf English} & \multicolumn{2}{c}{\bf Russian} \\
                        & Nouns        & Verbs        & Nouns        & Verbs        \\ \midrule
                        & \multicolumn{4}{c}{\bf fastText}                              \\  \midrule
Baseline           & 0.325        & 0.183     & 0.421        & 0.334                \\
Ranking          & 0.375        & 0.190          & 0.507        & 0.336              \\
Ranking + Wiki  & \textbf{0.400}        & \textbf{0.238}     & 0.540        & 0.383              \\ \midrule
                        & \multicolumn{4}{c}{\bf BERT}                                  \\ \midrule
Baseline           & 0.239        & 0.097      & 0.138        & 0.119               \\
Ranking        & 0.238        & 0.105            & 0.185        & 0.127              \\
Ranking + Wiki   & 0.253        & 0.120    & 0.218        & 0.161              \\ \midrule
                        & \multicolumn{4}{c}{\textbf{RUSSE'2020} participating systems}               \\ \midrule
Top-1 for Nouns: \textit{Yuriy}           & 0.328          & 0.230           & \textbf{0.552}        & 0.436                \\
Top-1 for Nouns: \textit{Yuriy}, no search engine features           & 0.300           & 0.231           & 0.507        & 0.388                \\
Top-1 for Verbs: \cite{dale2020russe}        & 0.234           & 0.224      & 0.418        & \textbf{0.448}                \\ \bottomrule
\end{tabular}
\caption{MAP scores for the taxonomy enrichment methods for English (2.0-3.0) and Russian datasets.}
\label{tab:results_rus_en}
\end{table}


\subsection{Results}

We test the models suggested in Section \ref{sec:baselines} on our new created English and Russian datasets as well as on the RUSSE'2020 dataset. Table \ref{tab:results_rus_en} compares the performance of our models with fastText and BERT embeddings on RUSSE'2020 and the symmetrical  subset of the English dataset where named entities and short words are excluded. 
Table \ref{tab:results_groups} demonstrates the results for the non-restricted datasets with the best-performed embeddings. Table \ref{tab:examples} illustrates top-10 candidates generated by the best performing system. Due to size limitation, we report the results only for \textit{WordNet 2.0-3.0} dataset, but the performance of all models on the other English datasets is consistent with that on this corpus.

\begin{table}[]
\centering
\begin{tabular}{r|r|r|r|r}
\toprule
\multicolumn{1}{c|}{\textbf{Rank}}                                                                & \multicolumn{1}{c|}{\textbf{dancing-master}}                            & \multicolumn{1}{c|}{\textbf{(to) ooh}}                                                                              & \multicolumn{1}{c|}{\textbf{Cinderella}}                                       & \multicolumn{1}{c}{\textbf{(to) go cheap}}                                    \\
\midrule
1                                                                                                & {\color[HTML]{009901} { \textbf{dancer.n.01}}}                      & {\color[HTML]{009901} {\textbf{exclaim.v.01}}}                                                            & {\color[HTML]{009901} { \textbf{mythical person}}}                         & to overdo                                                                \\
2                                                                                                & {\color[HTML]{009901} { \textbf{educator.n.01}}}                    & utter.v.02                                                                                                    & narrative  prose                                                              & to price                                                                 \\
3                                                                                                & {\color[HTML]{009901} { \textbf{performer.n.01}}}                   & sound.v.02                                                                                                    & wizard, magician, sorcerer                                                    & to buy                                                                   \\
4                                                                                                & {\color[HTML]{009901} { \textbf{teacher.n.01}}}                     & breathe.v.01                                                                                                  & sorceress                                                                     & to overestimate                                                          \\
5                                                                                                & ballet\_dancer.n.01                                                    & tremble.v.01                                                                                                  & human being                                                                   & to lower                                                                 \\
6                                                                                                & attendant.n.01                                                         & murmur.v.01                                                                                                   & {\color[HTML]{009901} { \textbf{fairy tale}}}                              & to increase                                                              \\
7                                                                                                & orator.n.01                                                            & impress.v.02                                                                                                  & nobleman                                                                      & to pay                                                                   \\
8                                                                                                & schoolteacher.n.01                                                     & shout.v.02                                                                                                    & short story                                                                   & {\color[HTML]{009901} { \textbf{to sell}}}                            \\
9                                                                                                & chaperon.n.01                                                          & talk.v.02                                                                                                     & female, woman                                                                 & to overcharge                                                            \\
10                                                                                               & principal.n.02                                                         & {\color[HTML]{009901} { \textbf{express.v.02}}}                                                            & poor person                                                                   & {\color[HTML]{009901} { \textbf{to act}}}                             \\
\midrule\midrule
\multicolumn{1}{c|}{}                                                                             & \begin{tabular}[c]{@{}r@{}}dancer.n.01, \\ performer.n.01\end{tabular} & \multicolumn{1}{c|}{}                                                                                          & \begin{tabular}[c]{@{}r@{}}fairy tale, \\  literary fairy tale\end{tabular}   & \begin{tabular}[c]{@{}r@{}}to sell, \\ to deliver posession\end{tabular} \\ \cline{2-2} \cline{4-5} 
\multicolumn{1}{c|}{\multirow{-2}{*}{\begin{tabular}[c]{@{}c@{}}\textbf{Ground} \\ \textbf{truth}\end{tabular}}} & \begin{tabular}[c]{@{}r@{}}educator.n.01, \\ teacher.n.01\end{tabular} & \multicolumn{1}{c|}{\multirow{-2}{*}{\begin{tabular}[c]{@{}l@{}}express.v.02,\\ exclaim.v.01\end{tabular}}} & \begin{tabular}[c]{@{}r@{}}imaginary creature,\\ mythical person\end{tabular} & \begin{tabular}[c]{@{}r@{}}to make a mistake,\\ to act\end{tabular}    \\
\bottomrule
\end{tabular}
\caption{Predicted hypernym synsets from
WordNet (\textit{dancing-master, ooh}) and RuWordNet (\textit{Cinderella, go cheap}). Underlined green bold text denotes predictions
of the model from the ground truth.}
\label{tab:examples}
\end{table}

\begin{table}[ht!]
\centering
\begin{tabular}{l|lll|l|ll|l}
\toprule
\multicolumn{1}{c|}{\multirow{2}{*}{\bf Method}} & \multicolumn{4}{c|}{\bf Nouns}             & \multicolumn{3}{c}{\bf Verbs}             \\ \cmidrule{2-8}
\multicolumn{1}{c|}{}                        & NE     & Short & Other & All    & Short & Other & All   \\ \midrule
\% in the data                              & 38\%  & 6\% & 61\%  & --       & 5\%   & 95\%  & --       \\ \midrule
              & \multicolumn{7}{c}{WordNet 2.0 --- WordNet 3.0 (fastText)}                                                 \\ \midrule
Baseline      & 0.328 & 0.319 & 0.233 & 0.291 &  \textbf{0.444} & 0.191 & 0.205 \\
Ranking        & 0.424 & 0.348 & 0.288 & 0.339 & 0.296 & 0.208 & 0.213 \\
Ranking + Wiki & \textbf{0.437}  & \textbf{0.360} & \textbf{0.332} & \textbf{0.372}  & 0.411 & \textbf{0.263} & \textbf{0.271} \\ \midrule
              & \multicolumn{7}{c}{RuWordNet 1.0 --- RuWordNet 2.0 (fastText)}                                                     \\ \midrule
\% in the data & 25\%   & 7\% & 70.7\%  & --      &  0\%  & 100\%  & --       \\ \midrule
Baseline      & 0.251 & 0.165 & 0.337 & 0.309 & -  & 0.232 & 0.232 \\
Ranking        & 0.417 & 0.218 & 0.381 & 0.384 & -  & 0.252 & 0.252 \\
Ranking + Wiki & \textbf{0.436} & \textbf{0.230} & \textbf{0.416} & \textbf{0.414} & -  & \textbf{0.295} & \textbf{0.295} \\ \bottomrule
\end{tabular}
    \caption{MAP scores for different categories of words for English and Russian datasets.}
    \label{tab:results_groups}
\end{table}


From Tables \ref{tab:results_rus_en} and \ref{tab:results_groups} we see that our methods consistently improve the hypernym detection for both nouns and verbs across different datasets. Extending a list of hypernym candidates with second-order hypernyms and ranking them increases MAP by a large margin, especially for nouns. Adding Wiktionary features further boosts the performance of models. 
However, the use of contextualised word embeddings demonstrated in Table \ref{tab:results_rus_en} does not guarantee high results in this task. The models which used BERT vector representations perform worse than the same approaches using fastText. This also holds for all datasets and parts of speech. Apparently, fastText is good at modelling common and the most popular word senses, whereas BERT embeddings aggregate sense from different contexts, which results in mixing different senses and confusing word representations. For example, for the word ``\cyrins{смайлик}'' (emoji) the predicted candidates are completely unsuitable (person, device, flatterer, hypocrite, visual materials) in comparison with the fastText prediction and correct hypernyms (graphical sign, image, symbol). Therefore, the contexts (in a broad sense) extracted from the pre-trained fastText embeddings are sufficient to attach new words to the taxonomy.



Our methods were not able to outperform the best-performing systems from RUSSE'2020 shared task for Russian. However, as it has been discussed in Section \ref{sec:related}, the best approach for nouns relies on external sources which are difficult to reproduce. In contrast to that, our approach is based on pure fastText vectors, word similarities, and Wiktionary which is available for multiple languages. At the same time, the approach by \newcite{dale2020russe} was ranked first in the verbs track. It does not use additional sources, but it only suits for Russian verbs, because it performed below the baseline at Russian nouns and at the whole English dataset. Our method suits for both verbs and nouns and is stable across languages.



\subsection{Error Analysis}

To better understand the difference in systems performance and their main difficulties, we made a quantitative and qualitative analysis of the results.

\subsubsection{Performance on Different Classes of Words}

We noticed that for certain words hypernym discovery is an easier task. In particular, named entities and some other categories of words seem less challenging for our models. To test that, we divide our datasets into several parts: named entities, short words (less than 4 letters) and the rest. We compute MAP separately for each of these groups (see Table~\ref{tab:results_groups}).

The MAP scores vary significantly across groups. Since MAP for a dataset is an average of MAPs for individual words, we can directly compare scores for different subsets.
Thus, we see that for both languages named entities are easier to find hypernyms for. This happens because their hypernyms often contain the word from the same named entity. For instance, the named entity ``Massif Central'' has ``massif.n.01'' as one of the true hypernyms. 
The performance on short nouns and short verbs also differs. Whereas short nouns are often polysemous (hence more challenging), short verbs have one sense, are uncommon, and their sense is sometimes deduced from their form (e.g. ``to aah'' --- ``to produce an `aah' sound'').

Finally, the performance on all other nouns and verbs which have no such lexical cues is lower than on the whole list of words. This trend is particularly marked for nouns where less challenging groups (NE) constitute less than two fifths in both datasets. Thus, in order to evaluate taxonomy enrichment models, we should check their quality on different groups of words. 

\begin{figure}[ht]
    \centering
    \subfloat[Russian dataset (nouns and verbs)]{{\includegraphics[width=0.5\textwidth]{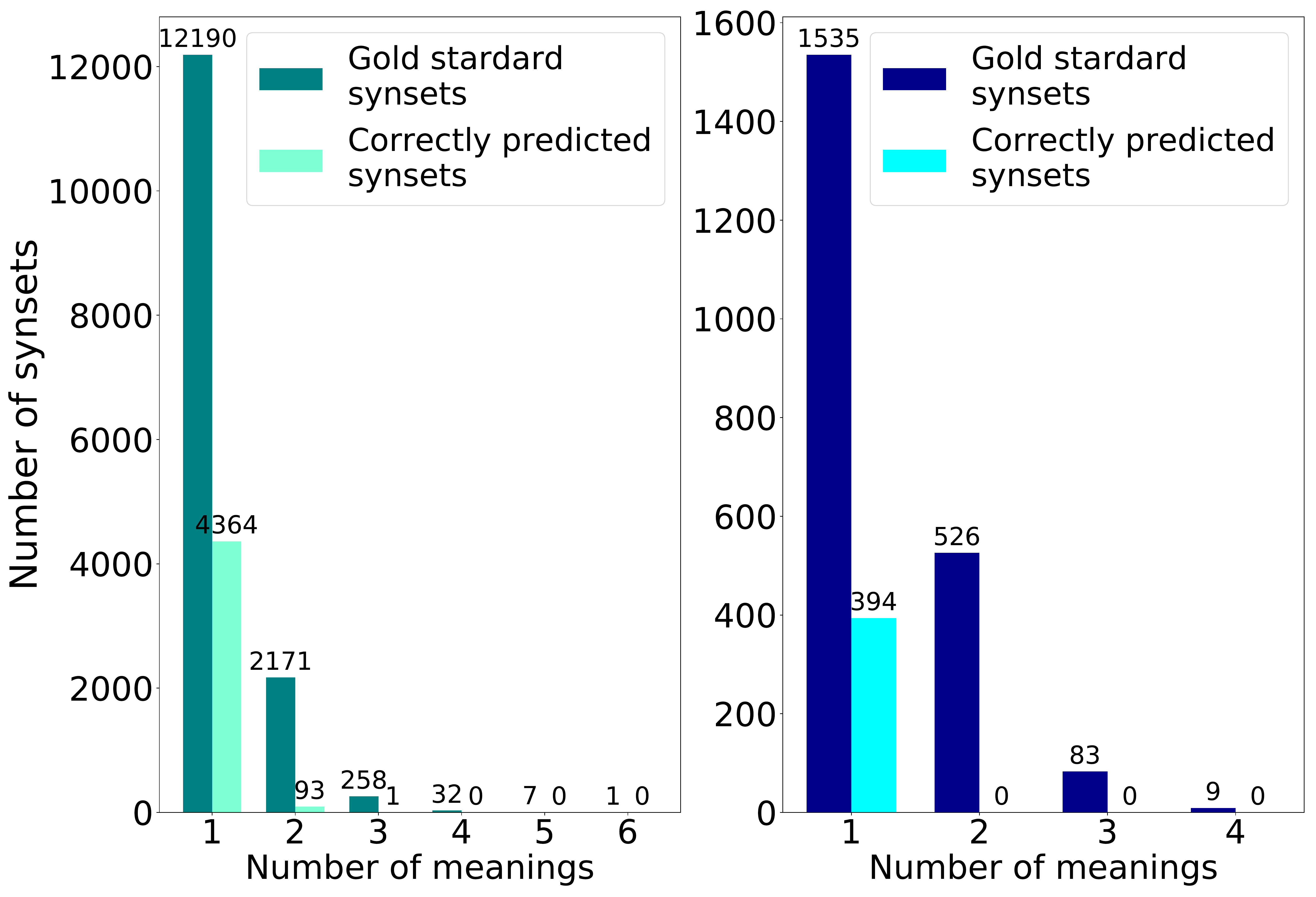} }}%
    \subfloat[English dataset (nouns and verbs)]{{\includegraphics[width=0.5\textwidth]{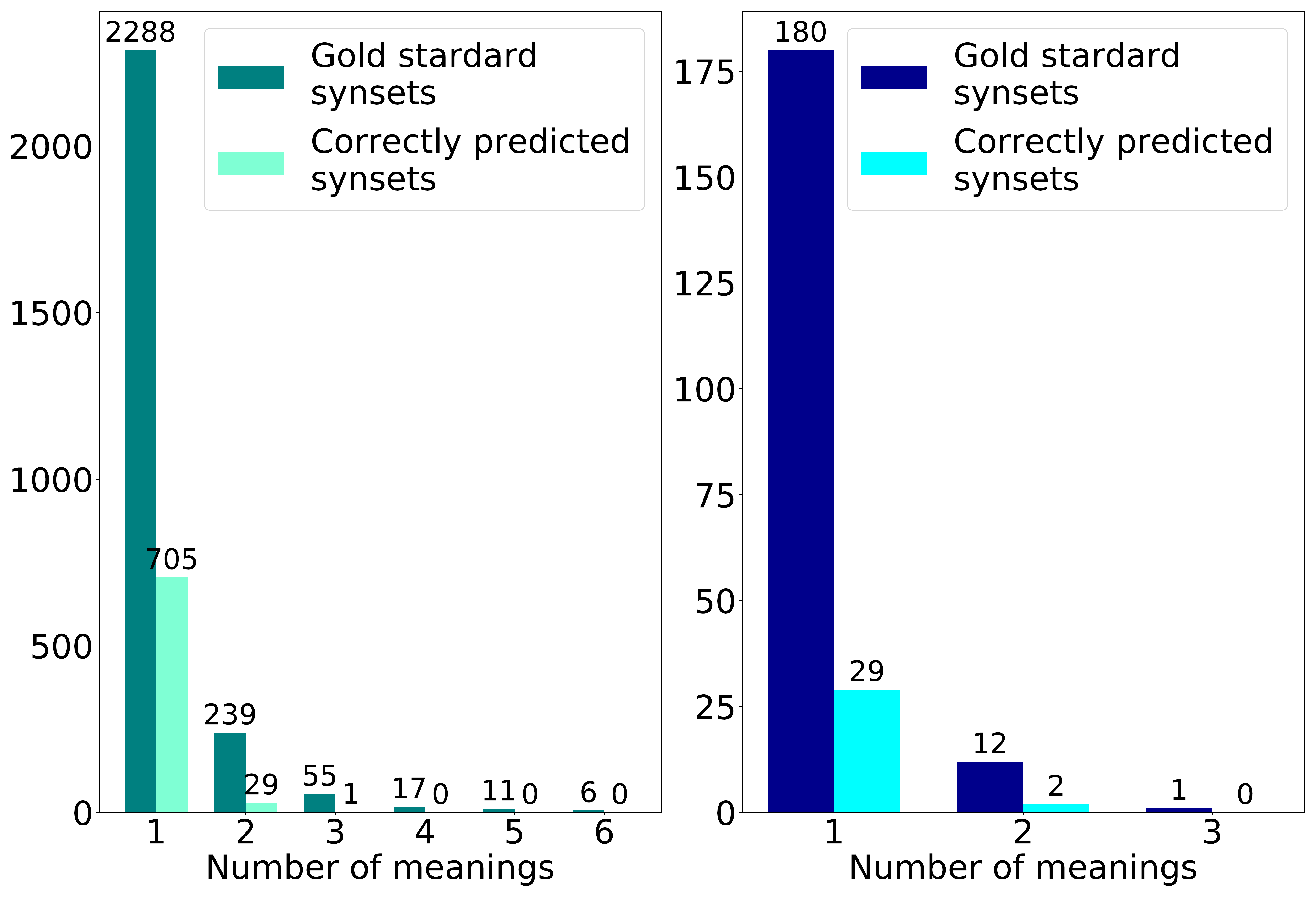} }}%
    \caption{Distribution of words over the number of senses.}
    \label{fig:distributions}
\end{figure}


\subsubsection{Distribution of Scores}




The differences in word semantics make the dataset uneven. In addition to that, we would also like to understand whether the performance of models depends on the number of connected components (possible meanings) for each word. Thus, we examine how many words with more than one meaning can be predicted by the system.

Figure \ref{fig:distributions}
depicts the distribution of synsets over the number of senses they convey. As we can see, the vast majority of words are monosemous. For Russian nouns, the system correctly identifies almost half of them, whereas for other datasets the share of correctly predicted monosemous words is below 30\%. This stems from the fact that for distributional models it is difficult to capture multiple senses in one vector. They usually capture the most widespread sense of a word. Therefore, the number of predicted synsets with two or more senses is extremely low. A similar power law distribution would be obtained using BERT embeddings, as we are still averaging embeddings from all contexts. This may be one of the reasons why contextualised models did not perform better than the fastest models which capture the main meaning only but do it well.

\subsubsection{Error Types}

In order to understand why a large number of word hypernyms (at least 60\%) are too difficult for models to predict, we turn to manual analysis of the system outputs. We find out that errors can be divided into two groups: system errors caused by distributional models limitations and taxonomy inaccuracies.
Therefore, we come across five main error types:

{\textbf{Type 1.} Extracted nearest neighbours can be semantically related words but not necessary co-hyponyms:}
    \begin{itemize}[noitemsep]
        \item delist (WordNet); expected senses: get rid of; predicted senses: remove, delete;
        \item \cyrins{хэштег} (hashtag, RuWordNet); expected senses: \cyrins{отличительный знак, пометка} (tag, label); predicted senses: \cyrins{символ, короткий текст} (symbol, short text).
    \end{itemize}
    
{\textbf{Type 2.} Distributional models are unable to predict multiple senses for one word:}
    \begin{itemize}[noitemsep]
        \item latakia (WordNet); expected senses: tobacco; municipality city; port, geographical point; predicted senses: tobacco;
        \item \cyrins{запорожец} (zaporozhets, RuWordNet); expected senses: \cyrins{житель города} (citizen, resident); \cyrins{марка автомобиля, автомобиль} (car brand, car); predicted senses: \cyrins{автомобиль, мототранспортное средство, марка автомобиля} (car, motor car, car brand).
    \end{itemize}
    
\textbf{Type 3.} System predicts too broad / too narrow concepts:
    \begin{itemize}[noitemsep]
        \item midweek (WordNet); expected senses: day of the week, weekday; predicted senses: time period, week, day, season;
        
        
        \item \cyrins{медянка} (smooth snake, RuWordNet); expected senses: \cyrins{неядовитая змея, уж} (non-venomous snake, grass snake); predicted senses: \cyrins{змея, рептилия, животное} (snake, reptile, animal).
        

    \end{itemize}
    
\textbf{Type 4.} Incorrect word vector representation: nearest neighbours are not semantically close:
    \begin{itemize}[noitemsep]
        \item falanga (WordNet); expected senses: persecution, torture; predicted senses: fish, bean, tree, wood.;
        
        \item \cyrins{кубокилометр} (cubic kilometer, RuWordNet); expected senses: \cyrins{единица объема, единица измерения} (unit of capacity, unit of measurement); predicted senses: \cyrins{город, городское поселение, кубковое соревнование, спортивное соревнование} (city, settlement, competition, sports contest).

    \end{itemize}
    
\textbf{Type 5.} Unaccounted senses in the gold standard datasets, inaccuracies in the manual annotation:
    \begin{itemize}[noitemsep]
        \item emeritus (WordNet); expected senses: retiree, non-worker; predicted senses: professor, academician;
        \item \cyrins{сепия} (sepia, RuWordNet); expected senses: \cyrins{морской моллюск} ``sea mollusc''; predicted senses: \cyrins{цвет, краситель} (color, dye).

    \end{itemize}



The above mentioned mistakes and inaccuracies may dramatically decrease the scores of automatic metrics. 
In order to check how useful the predicted synsets are for a human annotator (i.e. if a short list of possible hypernyms can speed up the manual extension of a taxonomy), we conduct the manual evaluation of 10 random nouns and 10 random verbs for both languages (the words are listed in Table \ref{tab:top20}). We focus on worse-quality cases and thus select words whose MAP score is below 1. Annotators with the expertise in the field and the knowledge of English and Russian were provided with guidelines and asked to evaluate the outputs from our best-performing system. Each word was labelled by 4 expert annotators, Fleiss's kappa is $0.63$ (substantial agreement) for both datasets.

\begin{figure}%
    \centering
    \subfloat[Russian dataset]{{\includegraphics[width=0.5\textwidth]{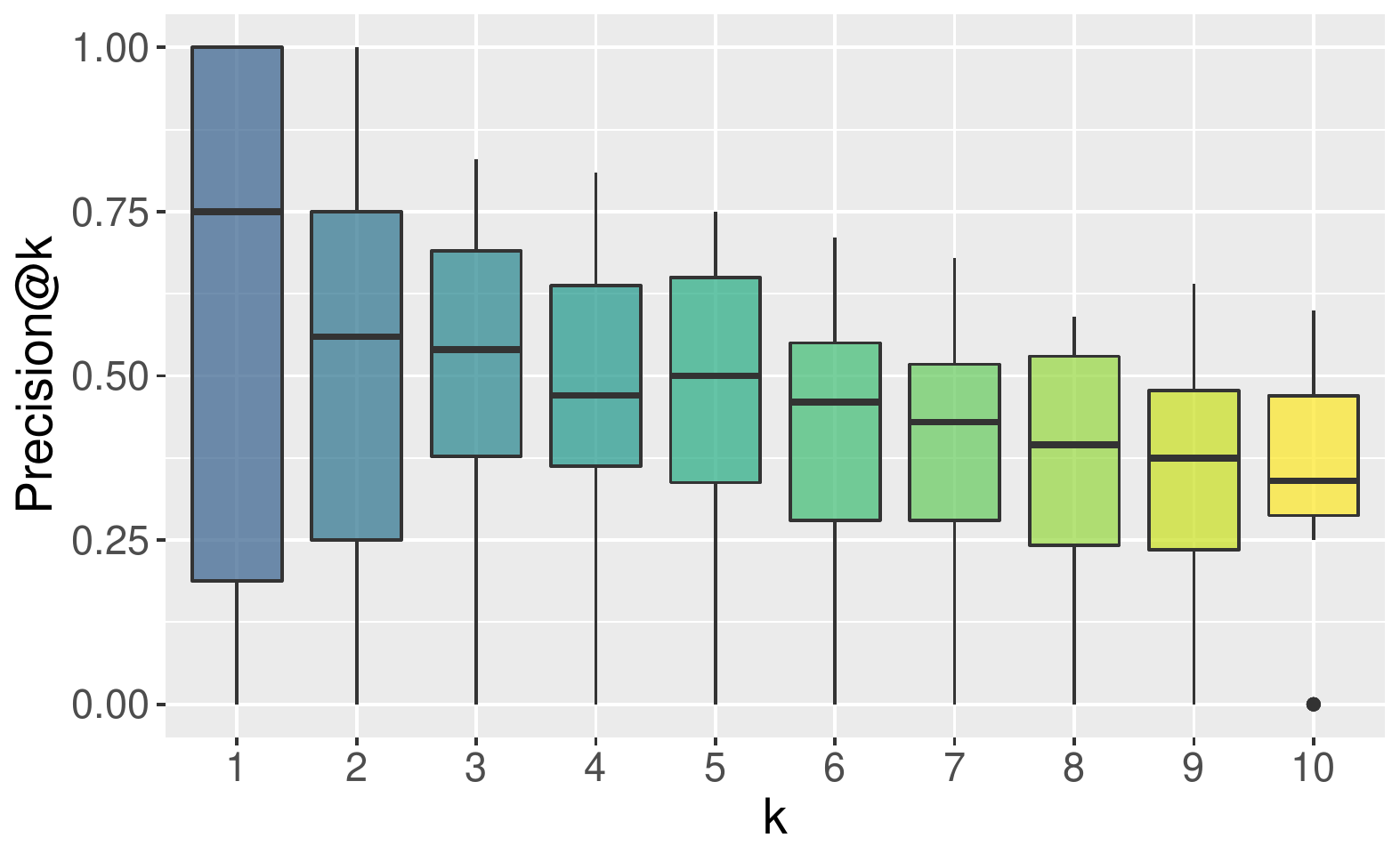} }}%
    \subfloat[English dataset]{{\includegraphics[width=0.5\textwidth]{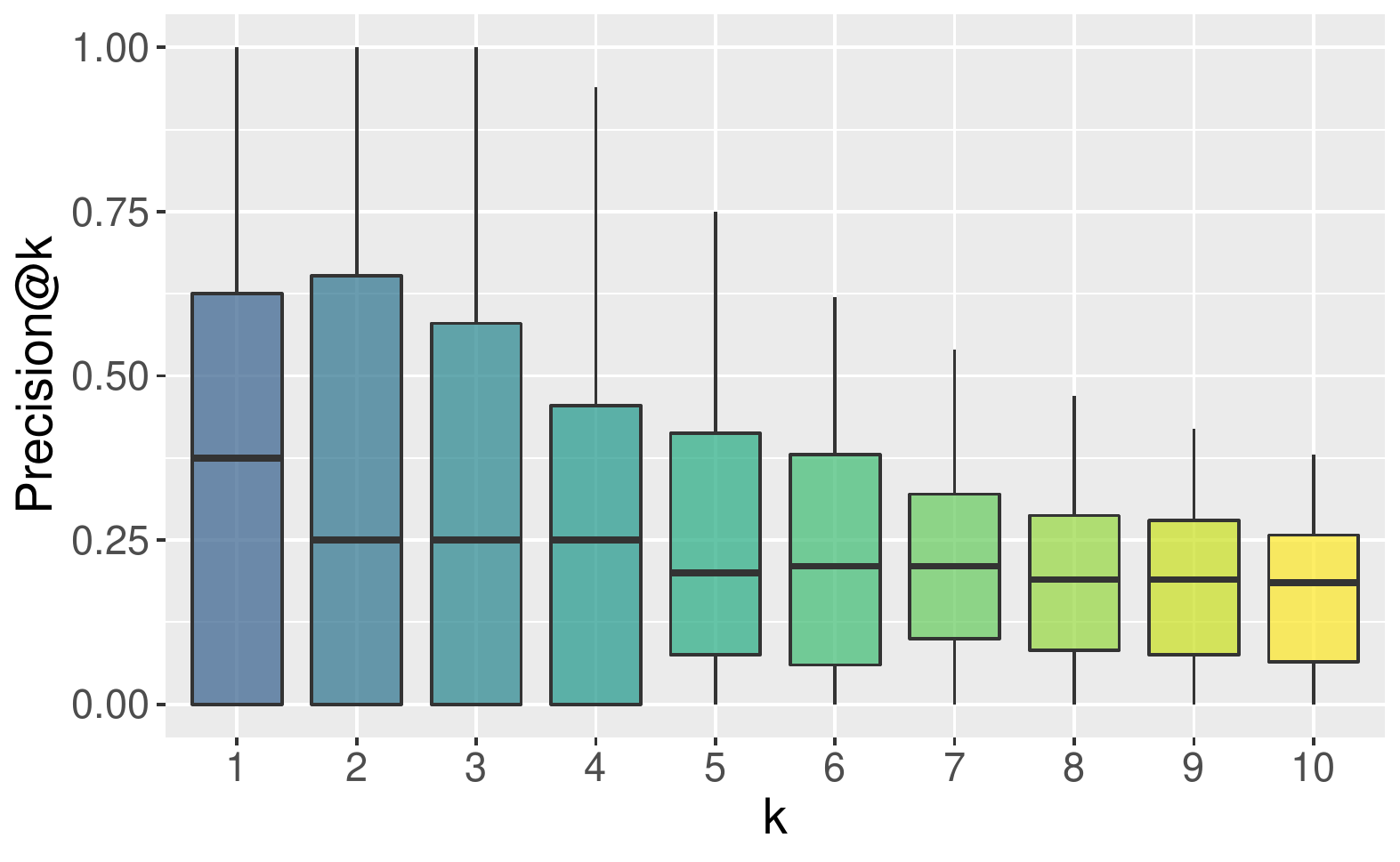} }}%
    \caption{Manual datasets evaluation results: Precision@10.}%
    \label{fig:precision}%
\end{figure}

We compute Precision@k score (the share of correct answers in the generated lists from position 1 to $k$) for $k$ from 1 to 10, shown in Figure \ref{fig:precision}. We can see that even for words with MAP below 1 our model manages to extract useful hypernyms. 

\begin{table}[]
\centering
\resizebox{\textwidth}{!}{
\begin{tabular}{cl}
\toprule
\textbf{Language} & \textbf{Word List} \\
\midrule
English & \begin{tabular}[c]{@{}l@{}}falanga, venerability, ambulatory, emeritus, salutatory address, eigenvalue of a matrix, \\ liposuction, moppet, dinette, snoek, to fancify, to google, to expense, to porcelainize, \\ to junketeer,  to delist, to podcast, to deglaze, to shoetree, to headquarter\end{tabular} \\
\midrule
Russian & \begin{tabular}[c]{@{}l@{}}\cyrins{барабашка}, \cyrins{листинг}, \cyrins{стихосложение}, \cyrins{аукционист}, \cyrins{точилка},  \cyrins{гиперреализм},\\ \cyrins{серология},  \cyrins{огрызок},  \cyrins{фен},  \cyrins{марикультура},  \cyrins{уломать},  \cyrins{отфотошопить}, \cyrins{тяпнуть}, \\ \cyrins{растушевать},  \cyrins{завраться},  \cyrins{леветь},  \cyrins{мозолить},  \cyrins{загоститься},  \cyrins{распеваться},    \cyrins{оплавить} 
\end{tabular}       \\
\bottomrule
\end{tabular}}
\caption{Words selected for the manual evaluation.}
\label{tab:top20}
\end{table}



\section{Conclusion}




In this work, we deal with the taxonomy enrichment task: the automatic extension of an existing taxonomy with new terms. The novelty of our work is the use of diachronic versions of wordnets in two languages: this setting reflects the real process of development of the resources in time and allows us to study if machines could perform similar taxonomy completion task automatically. Toward this end, we present a simple method for solving the task that can be 
applied to multiple languages. Its results are 
close to the state-of-the-art for Russian and stable across languages. An interesting finding is that the task does not benefit from context-informed embeddings,
whereas 
context-free vector representations like fastText 
often successfully identify hypernyms. 
On the other hand, the availability of Wiktionary substantially boosts the performance, yet 
models based solely on embeddings 
still yield competitive results. 

Error analysis reveals that some groups of words (e.g. named entities) are easier to find a hypernym for, and polysemous words are significantly more challenging. At the same time, our models were able to identify some cases of polysemy which were not reflected in wordnets.
Promising directions of future work are (i) application of the developed methods to automate work of lexicographers constructing wordnets and (ii) further improvement of results for verbs.

\section*{Acknowledgments}
The work of Natalia Loukachevitch in the current study (preparation of RuWordNet data for the experiments)  is supported by the Russian Science Foundation (project 20-11-20166). We thank Yuriy Nazarov and David Dale for running their approaches from RUSSE'2020 shared task on the English datasets.

\bibliographystyle{coling}
\bibliography{coling2020}

\end{document}